\documentclass[conference]{IEEEtran}
\IEEEoverridecommandlockouts
\usepackage{cite}
\usepackage{amsmath,amssymb,amsfonts}
\usepackage{algorithmic}
\usepackage{graphicx}
\usepackage{textcomp}
\usepackage{booktabs}
\usepackage{multirow}
\usepackage{url}
\usepackage{comment}
\usepackage{tikz}
\usepackage{amsmath}  
\usetikzlibrary{arrows.meta, positioning, calc}
\usepackage{flushend}
\def\BibTeX{{\rm B\kern-.05em{\sc i\kern-.025em b}\kern-.08em
    T\kern-.1667em\lower.7ex\hbox{E}\kern-.125emX}}
\begin{document}

\title{Machine Learning-based sEMG Signal Classification for Hand Gesture Recognition}



\author{
\IEEEauthorblockN{Parshuram N. Aarotale}
\IEEEauthorblockA{\textit{Dept. of Biomedical Engineering} \\
\textit{Wichita State University, USA} \\
pnaarotale@shockers.wichita.edu}
\and

\IEEEauthorblockN{Ajita Rattani}
\IEEEauthorblockA{\textit{Dept. of Computer Science and Engineering} \\
\textit{Uni. of North Texas, USA} \\
ajita.rattani@unt.edu}}

\maketitle

\begin{abstract}
EMG-based hand gesture recognition uses electromyographic~(EMG) signals to interpret and classify hand movements by analyzing electrical activity generated by muscle contractions. It has wide applications in prosthesis control, rehabilitation training, and human-computer interaction. Using electrodes placed on the skin, the EMG sensor captures muscle signals, which are processed and filtered to reduce noise. 
Numerous feature extraction and machine learning algorithms have been proposed to extract and classify muscle signals to distinguish between various hand gestures.

This paper aims to benchmark the performance of EMG-based hand gesture recognition using novel feature extraction methods, namely, fused time-domain descriptors, temporal-spatial descriptors, and wavelet transform-based features, combined with the state-of-the-art machine and deep learning models. 
Experimental investigations on the Grabmyo dataset demonstrate that the 1D Dilated CNN performed the best with an accuracy of $97\%$ using fused time-domain descriptors such as power spectral moments, sparsity, irregularity factor and waveform length ratio. Similarly, on the FORS-EMG dataset, random forest performed the best with an accuracy of $94.95\%$ using temporal-spatial descriptors (which include time domain features along with additional features such as coefficient of variation (COV), and Teager-Kaiser energy operator (TKEO)).

\end{abstract}


\section{Introduction}
\label{introduction}

Surface electromyogram~(sEMG) signals are pivotal in the development of effective human-machine interface~(HMI). These systems are deployed in a wide range of applications, including the control of myoelectric prostheses~\cite{atzori2016deep}, providing rehabilitative feedback~\cite{kumar2019prosthetic}, predicting diseases~\cite{muzammal2020multi}, and advancing neurorobotics~\cite{gulati2021toward}. Further, the use of sEMG for gesture recognition in assistive technologies is becoming increasingly prevalent among individuals with limb amputations. For example, myoelectric prostheses use sEMG signals to interpret muscle contractions and generate corresponding hand movements, thus improving functionality and user experience with prosthetic limbs.

Recent advances in hand gesture recognition using surface electromyographic (sEMG) signals have explored both traditional and deep learning methods to improve accuracy and control. Methods such as k-Nearest Neighbor (KNN), Linear Discriminant Analysis (LDA), and Support Vector Machine (SVM) have been successfully applied to classify hand-gesture movements ~\cite{khushaba2012toward,tuncer2022novel,khushaba2012electromyogram,essa2023features,pancholi2018portable}. Hybrid approaches that combine fuzzy cognitive maps (FCM) and Bayesian belief networks with extreme learning machines (ELM) have also been investigated, demonstrating their effectiveness in hand gesture classification ~\cite{prabhakar2023efficient}. Feature extraction techniques such as variational mode decomposition (VMD) within a multiclass SVM framework showed high precision in hand gesture classification~\cite{prabhavathy2024hand}. The fractional Fourier transform (FrFT) features have been used with KNN to classify the movements of the ten fingers~\cite{taghizadeh2021finger}, obtaining competitive results. 

Deep learning techniques have been utilized for real-time hand gesture recognition, obtaining an accuracy of $94\%$ and often outperforming traditional algorithms~\cite{lee2021electromyogram}. Convolutional neural networks (CNN) have been widely used for hand gesture recognition ~\cite{asif2020performance,chen2020hand,wang2023hand,zhang2022emg,wang2023emg}. Additionally, several studies have used Long Short Term Memory (LSTM) networks for hand gesture recognition obtaining comparable performance as that of CNN~\cite{ovur2021novel,barron2020recurrent,topalovic2019emg}.

Although existing studies in this field have obtained several encouraging findings using machine and deep-learning algorithms, it is difficult to analyze their practical implications due to the restricted settings and datasets used in these studies~\cite{pancholi2018portable,zhang2022emg,wang2023emg,chamberland2023novel,ghislieri2021long}.
This indicates that these studies utilize datasets gathered from various sensors, which were deployed at different sampling frequencies and positioned at diverse body locations such as elbow and forearm. This variability can impact the consistency and comparability of the findings across the studies.

In order to better understand and compare the performance, this paper \textbf{aims} to benchmark the performance of the novel feature extraction methods along with several machine and deep-learning classifiers for hand gesture recognition.

\vspace{0.25 cm} \noindent\textbf{Our Contributions:} The contributions of this paper are as follows:
\begin{itemize}

\item \textbf{Robust Feature Extraction:} This paper introduces novel feature extraction techniques such as (time domain along with temporal-spatial domain) power spectral moments, sparsity, irregularity factor (IRF), waveform length ratio (WLR), coefficient of variation (COV), and Teager-Kaiser energy operator (TKEO) and wavelet transform-based features such as Energy, Variance, Standard Deviation, Waveform Length, and Entropy for hand gesture recognition.

\item \textbf{Benchmarking Machine and Deep Learning Models:} This paper benchmarked multiple machine learning algorithms such as LDA, SVM, KNN, Random forest, ensemble learning, and deep learning algorithms, namely 1D Dilated CNN and 1D Dilated CNN-LSTM in combination with the aforementioned feature extraction techniques for hand gesture recognition.

\item \textbf{Experimental Validation on Latest Datasets:} Experimental investigation on two publicly available latest~(2024) datasets, namely, Grabmyo~\cite{pradhan2022multi} and FORS-EMG~\cite{rumman2024fors}. 

\end{itemize}

This paper is organized as follows. In section $2$, we discuss the related work on EMG-based hand gesture recognition using machine and deep learning models. In section $3$, we discuss the feature descriptors used in this study. In Section $4$, the datasets used along with the pre-processing techniques, implementation details, and evaluation metrics are discussed. The results are analyzed in section $5$ along with the discussion. Conclusions and future directions are discussed in Section $6$.

\section{Related Work}

Recent advances in surface electromyography (sEMG) for hand gesture recognition includes various techniques that employ both traditional machine learning and advanced deep learning models. This section reviews the contributions of several studies in this domain.

\subsection{Traditional Machine Learning Approaches}
Tuncer et al.~\cite{tuncer2022novel} employed a multi-level feature extraction technique and fine-tuned k-Nearest Neighbor (KNN) and Support Vector Machine (SVM) classifiers to classify $15$ individual and combined finger movements were evaluated on the public dataset~\cite{khushaba2012electromyogram}. Provakar et al.~\cite{prabhakar2023efficient} proposed four hybrid models for finger movement classification based on graph entropy, fuzzy cognitive maps (FCM), empirical wavelet transformation (EWT), a fuzzy clustering technique combined with a least squares support vector machine classifier (LS-SVM), and Bayesian belief networks (BBN) with extreme learning machines (ELM). 
Essa et al.~\cite{essa2023features} evaluated five feature extraction techniques using k-nearest Neighbor (KNN), Linear Discriminant Analysis (LDA) and Support Vector Machine (SVM) to categorize $17$ gestures with the LDA classifier demonstrating substantial classification accuracy. 

Lee et al.~\cite{lee2021electromyogram} developed a real-time gesture recognition system for hand and finger movements using $18$ time-domain features. The study found that Artificial Neural Networks (ANN) outperformed other algorithms (SVM, Random forest, and Logistic Regression) with a $94\%$ accuracy rate.

\subsection{Deep Learning Approaches}

Convolutional Neural Networks (CNNs) have been widely employed to address the hand gesture recognition problem in numerous studies in the literature~\cite{wang2023hand,zhang2022emg,wang2023emg}. For instance, Atzori et al.~\cite{atzori2016deep} classified approximately $50$ gestures from one of the NinaPro databases using a basic CNN architecture. 
Yang et al.~\cite{yang2019emg} explored the use of raw EMG signal data as input to CNNs, comparing time-domain and frequency-domain representations across two publicly available datasets, obtaining favorable results. 

Chamberland et al.~\cite{chamberland2023novel} developed EMaGer, a $64$-channel HD-EMG sensor that is expandable and adaptable, capable of fitting various forearm sizes, and utilizes a CNN-based model for gesture classification.
Additionally, LSTM networks have been utilized in various studies. For example, Ghislieri et al.~\cite{ghislieri2021long} demonstrated that LSTM networks can accurately detect muscle activities in EMG recordings without cancelation of background noise. 
In contrast, Antonius et al.~\cite{antonius2021electromyography} obtained enhanced results by combining CNNs with an LSTM-like recurrent neural network, although their success was limited to a small set of basic gestures. 
López et al.~\cite{lopez2024cnn} demonstrated the effectiveness of a CNN-LSTM model for hand gesture recognition based on EMG signals. The CNN-LSTM model outperformed a CNN-only model, obtaining a recognition accuracy of $90.55$ ± $9.45$ after post-processing, compared to $87.26$ ± $11.14$ for the CNN-only model.

\section{Methodology: Feature Extraction Methods}

In this study, several feature extraction methods were applied such as wavelet transformation-based features, fused time-domain descriptors (fTDD) and temporal-spatial descriptors (TSD) to capture both temporal and spatial properties of electromyographic (EMG) signals for gesture recognition. These features were obtained using a sliding window approach in which the signals were divided into $600$ ms overlapping windows with a $50\%$ overlap between them.

\subsection{Fused Time-Domain Descriptors (fTDD)}

The fTDD approach was used on each channel separately to extract important features in the time domain that capture temporal fluctuations in EMG signals.
Six features were calculated from each window, including power spectral moments (M0, M2, M4), sparsity, irregularity factor (IRF), and waveform length ratio (WLR); which jointly reflect signal energy, variability, and complexity. These features are explained in the Table~\ref{tab:TDDf}~\cite{khushaba2016fusion}. 

These signals are then transformed non-linearly by taking the logarithm of the squared signal; features such as power spectral moments (M0, M2, M4), sparsity, irregularity factor (IRF), and waveform length ratio (WLR) were extracted from transformed signal.

Following extraction of features from both the original and transformed signals, a correlation-based analysis is carried out~\cite{khushaba2016fusion}.

\begin{table*}[ht]
\caption{Summary of Fused Time-Domain Descriptors (fTDD) and Temporal Spatial Descriptor~(TSD) features.}
\label{tab:TDDf}
\centering
\small  
\begin{tabular}{| p{4cm} | p{7cm} | p{7cm} |}
\hline
\textbf{Feature Name} & \textbf{Description} & \textbf{Formula} \\ \hline

Root Squared Zero Order Moment & The feature represents the signal energy. It is normalized by dividing by the median of the zero-order moments from all channels. & $\tilde{m}_0 = \sqrt{\frac{1}{N} \sum_{j=0}^{N-1} x[j]^2}$ \\ \hline

Root Squared Second Order Moment & This feature is derived from the second derivative of the signal. It captures spectral properties based on the second derivative of the time-domain signal using the Fourier transform. & $\tilde{m}_2 = \sqrt{\frac{1}{N} \sum_{j=0}^{N-1} (\Delta x[j])^2}$ \\ \hline

Root Squared Fourth Order Moment & The feature is derived from the fourth derivative of the signal, capturing higher-order spectral properties from the signal's fourth derivative. & $\tilde{m}_4 = \sqrt{\frac{1}{N} \sum_{j=0}^{N-1} (\Delta^2 x[j])^2}$ \\ \hline

Normalized Moments & All moments (zero, second, fourth) are normalized by dividing the moments raised to power \(k\) by the median factor \(\lambda\). & $ m_0 = \frac{\tilde{m}_0^k}{\lambda}, \quad m_2 = \frac{\tilde{m}_2^k}{\lambda}, \quad m_4 = \frac{\tilde{m}_4^k}{\lambda}$ \\ \hline

Sparseness (S) & This feature quantifies how much energy of a vector is packed into only a few components, measuring sparseness of the signal. & $f_4 = \log(S), \quad S = \left( \frac{m_0}{\sqrt{m_0 - m_2} \sqrt{m_0 - m_4}} \right)$ \\ \hline

Irregularity Factor (IF) & As a measure of regularity, this feature captures the ratio of the number of upward zero crossings (ZC) to the number of peaks (NP), using spectral moments. & $f_5 = \log \left( \frac{ZC}{NP} \right) = \log \left( \sqrt{\frac{m_2}{m_0 m_4}} \right)$ \\ \hline

Waveform Length Ratio (WL) &  WL feature is defined as the ratio of the waveform length of the first derivative to that of the second derivative.  & $ f_6 = \log \left( \frac{\sum_{i=0}^{N-1} |\Delta^2 x_i|}{\sum_{i=0}^{N-1} |\Delta x_i|} \right) $ \\
\hline

Coefficient of Variation (COV) & This feature represents the ratio of the standard deviation to the mean of the EMG signal, measuring signal dispersion. & $f_7 = \log(COV) = \log \left( \frac{\sqrt{\frac{1}{N-1} \sum_{j=0}^{N-1} (x[j] - \bar{x})^2}}{\frac{1}{N} \sum_{j=0}^{N-1} x[j]} \right)$ \\ \hline

Teager-Kaiser Energy Operator & This non-linear operator measures instantaneous energy changes of signals composed of a single time-varying frequency. The logarithm of the summation of the TKEO is used to represent the EMG energy. & $f_8 = \log(\Psi) = \log \left( \sum_{j=0}^{N-2} x^2[j] - x[j-1]x[j+1] \right)$ \\ \hline

\end{tabular}
\end{table*}

\subsection{Temporal-Spatial Descriptors~(TSD)}

Temporal-Spatial Descriptors~(TSD) are used to extract significant characteristics from electromyogram (EMG) signals; A windowing approach is used to process the signals, applying a sliding window of $600$ ms with a $50\%$ overlap between them. Both within-channel and between-channel features are extracted for every window.

 To capture inter-channel connections for between-channel features, this study computes pairwise differences across all possible combinations of two channels for each window and features such as power spectral moments (M0, M2, and M4), as well as additional characteristics such as sparsity, IRF, and coefficient of variation (COV). Moreover, the Teager-Kaiser energy operator (TKEO) was calculated. These features are explained in Table~\ref{tab:TDDf}~\cite{khushaba2016fusion,khushaba2017framework}.

For every channel, within-channel features, such as power spectral moments (M0, M2, and M4), as well as additional characteristics such as sparsity, IRF, coefficient of variation (COV) and Teager-Kaiser energy operator (TKEO) are calculated in addition to between-channel features. The extracted features from each window are then concatenated to form the complete set of Temporal-Spatial Descriptors (TSD)~\cite{khushaba2017framework}.

\subsection{Wavelet Transform-Based Features}

The Wavelet Transform (WT) feature extraction method involves decomposing the sEMG signals into multiple scales using the discrete wavelet transform (DWT)~\cite{khushaba2020mswt}. The steps for extracting wavelet-based features are as follows.

\subsubsection*{Wavelet Decomposition}

The signal is decomposed into multiple levels (\( J = 5 \)) using the Symlet wavelet (sym8). The DWT coefficients are calculated for each level \( j \).

\[
W_j(t) = \sum_{i=1}^{N} x_i \psi_{j,i}(t)
\]

where \( W_j(t) \) are the wavelet coefficients at level \( j \), \( x_i \) are the signal values, and \( \psi_{j,i}(t) \) are the wavelet basis functions.

These wavelet coefficients were used to extract the following features: Energy, Variance, Standard Deviation, Waveform Length, and Entropy. The detailed formulas for these characteristics are presented in Table ~\ref{tab:features} ~\cite{khushaba2020mswt}.

\begin{table}[ht]
\caption{Wavelet Transform based Features.}
    \centering
    \begin{tabular}{| p{3cm} | p{5cm} |}
        \hline
        \textbf{Feature} & \textbf{Formula} \\ \hline
        Energy ($E_j$) &
        $E_j = \sum_{i=1}^{N} W_j(i)^2$ \\ \hline

        Variance ($\sigma_j^2$) &
        $\sigma_j^2 = \frac{1}{N} \sum_{i=1}^{N} (W_j(i) - \bar{W_j})^2$ \\ \hline

        Standard Deviation ($\sigma_j$) &
        $\sigma_j = \sqrt{\sigma_j^2}$ \\ \hline

        Waveform Length ($WL_j$) &
        $WL_j = \sum_{i=1}^{N-1} |W_j(i+1) - W_j(i)|$ \\ \hline

        Entropy ($H_j$) &
        $H_j = -\sum_{i=1}^{N} W_j(i)^2 \log(W_j(i)^2 + c)$ \\ \hline
    \end{tabular}
    
    \label{tab:features}
\end{table}

\subsection{Machine and Deep-Learning Classifiers}

This study included many machine learning models, Linear Discriminant Analysis (LDA), Support Vector Machine (SVM), K-Nearest Neighbors (KNN), and Random Forest (RF). To improve classification performance, ensemble techniques such as Bagging with KNN, Bagging with SVM, and AdaBoost with Random Forest were also used. In this study, deep learning classifiers such as the 1D Dilated Convolutional Neural Network (CNN) and 1D CNN-LSTM were used.

Briefly, SVM is based on identifying the best hyperplane to divide data into distinct classes. It works well in high-dimensional spaces and can handle non-linear interactions with kernel functions~\cite{mujeeb2022automatic}. 
KNN is a nonparametric method for classifying new instances in the feature space based on their similarity to training examples~\cite{soon2008support}.
LDA effectively establishes decision boundaries between classes by maximizing the ratio of variance between classes to variance within classes using a linear feature combination, assuming normally distributed data with equal class covariances ~\cite{hastie2009elements}. RF is an ensemble learning method that builds numerous decision trees during training and returns the mode of the classes as the prediction (classification) or the average prediction (regression) of the individual trees~\cite{zhang2017learning}. Bagging is a group machine learning technique in which multiple subsets of the training dataset are created by random sampling with replacement~\cite{witten2005practical,subasi2022surface}. 
Boosting is another ensemble learning technique that focuses on improving model performance by combining multiple weak learners to create a strong learner sequentially~\cite{witten2005practical,subasi2022surface}.

1D CNN architecture consists of three convolutional layers with increasing dilation rates along with the set of two fully connected layers of size $128$ and $64$, along with the output layer. 1D Dilated CNN-LSTM consists of extracting spatial characteristics, the proposed 1D CNN-LSTM model combines three 1D convolutional layers with dilated convolutions and batch normalization followed by max-pooling. After that, a three-layer LSTM network with $256$ hidden units processes these features to identify temporal dependencies. Robust classification is obtained by passing the final LSTM output through fully connected layers along with dropout~($0.5$) layer.

\section{Experimental Details}

\subsection{Datasets}

\subsubsection{Grabmyo~\cite{pradhan2022multi}}
The Grabmyo~\cite{pradhan2022multi} dataset involves $43$ healthy participants ($23$ men and $20$ women) recruited from the University of Waterloo, with data collected on days $1$, $8$ and $29$. The participants had an average age of $26.35$ years (±2.89), and the average length of the forearm of $25.15$ cm (±1.74 cm). Exclusion criteria included muscle pain, skin allergies, or inability to complete all sessions. The study followed ethical guidelines, with participants giving their informed consent and the study was approved by the Office of Research Ethics of the University of Waterloo ($31,346$).
The data was collected with a sampling rate of $2048$ Hz and comprises $16$ gestures~refer, including lateral prehension, thumb adduction, various finger oppositions and extensions, wrist flexion and extension, forearm supination and pronation, hand open, and hand close. Gestures were performed in a randomized order with a ten-second rest between each. Each session included seven runs of $17$ gestures (including rest), totaling $119$ contractions per session. This protocol was repeated on day $8$ and day $29$. Electrode positions and lists of gesture used while recording sEMG data were shown in fig ~\ref{figdataset} A and B.

\subsubsection{FORS-EMG~\cite{rumman2024fors}}
\begin{figure*} [ht!]
    \centering
    \includegraphics[width=0.92\linewidth]{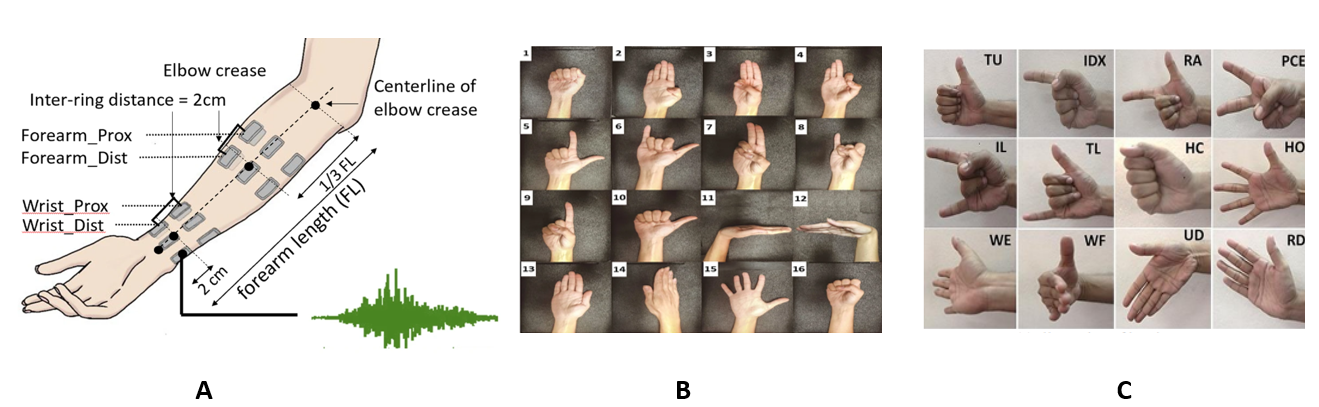}
    \caption{A) Electrode Positions~\cite{pradhan2022multi} and B) Gesture list for Grabmyo dataset~\cite{pradhan2022multi} and  C) FORS-EMG dataset~\cite{rumman2024fors}.}
    \label{figdataset}
\end{figure*}

The FORS-EMG dataset used in this work was collected from $19$ healthy individuals who performed $12$ different wrist and finger motions in three different forearm orientations: pronation, neutral (rest), and supination. The participants performed five repetitions of each gesture (see Fig ~\ref{figdataset} C) while the electrodes were positioned along the mid-forearm and close to the elbow. For each gesture, eight channels (four on the forearm and four around the elbow) were used to record sEMG signals at $8$ second intervals with a sampling frequency of $985$ Hz. This work processes only the dataset with orientation at rest.

\subsection{Preprocessing and implementation details}

Surface electromyography (sEMG) signals were bandpass filtered with a frequency of $20-450$Hz. Preprocessed signals were segmented into overlapping $0.6$-second windows with a $50\%$ overlap for feature extraction. The data sets were divided into $80\%$ for training and $20\%$ for testing, and the results were reported on the test set for all experiments.

Each machine learning classifier we used had its own set of hyperparameters. The Singular Value Decomposition (SVD) solver was employed in conjunction with Linear Discriminant Analysis (LDA). The Support Vector Machine (SVM) classifier was set up with enabled probability estimates, a linear kernel, and a regularization parameter of $1$. Using the Euclidean distance metric, K-Nearest Neighbors (KNN) classifier was used with $k$ equal to five neighbors. The ensemble of decision trees for the random forest was implemented using one hundred trees (estimators). Both KNN and SVM were used as base learners in the bagging process, and each bagging model was set up to employ $10$ base estimators. Lastly, AdaBoost was applied to Random Forest as a base learner. 

The deep learning models, namely, 1D dilated CNN and 1D dilated CNN-LSTM, were trained using the Adam optimizer with a learning rate of $0.0001$ and cross-entropy loss function. These models were trained using an early stopping mechanism.

\subsection{Evaluation Metrics}

The following standard performance metrics~\cite{fawcett2006introduction} were used
to evaluate the proposed models in this study.

\begin{align}
\text{Accuracy} &= \frac{T_{p} + T_{n}}{T_{p} + T_{n} + F_{p} + F_{n}} \\
\text{Precision} &= \frac{T_{p}}{T_{p} + F_{p}} \\
\text{Recall} &= \frac{T_{p}}{T_{p} + F_{n}} \\
F1\text{-score} &= 2 \times \frac{\text{Precision} \times \text{Recall}}{\text{Precision} + \text{Recall}}
\end{align}

where $T_{p}$, $T_{n}$, $F_{p}$, and $F_{n}$ represent true positive, true negative, false positive, and false negative recognition for the given class, respectively.

\section{Results and Discussion}

\subsection{Experimental Results on Grabmyo}


\begin{table}[ht]
\centering
\caption{Gesture recognition from time-domain descriptors using traditional machine learning and deep learning models tested on the Grabmyo dataset.}
\label{tab:TDDemg}
\resizebox{\columnwidth}{!}{%
\begin{tabular}{|l|c|c|c|c|}
\hline
\textbf{Models} & \textbf{ACC} & \textbf{P} & \textbf{R} & \textbf{F1} \\ \hline
LDA   & 84.26  &  0.85    & 0.84   &   0.84          \\
SVM   &   93.12 &   0.93   &  0.93  &  0.93         \\
KNN  &   93.86    &   0.94        &   0.94         &     0.94        \\
Random Forest  &   93.6  &   0.94   &   0.94   &   0.94        \\ 
Voting Ensemble  &       94.29      &   0.95          &   0.94         &     0.94        \\
Bagging KNN  &      94.33     &   0.94          &   0.94          &     0.94        \\
Bagging SVM  &     93.38     &   0.94          &   0.93          &     0.93        \\
Adaboost  &      93.61   &   0.94          &   0.94          &     0.94        \\
1D Dilated CNN  &      \textbf{97.00}    &   0.97          &   0.97         &     0.97        \\
1D Dilated CNN-LSTM  &      96.53   &   0.97          &   0.97         &     0.97        \\
\hline
\end{tabular}%
}
\end{table}

Table~\ref{tab:TDDemg} presents the performance metrics for various machine learning and deep learning models applied to gesture recognition using fused time domain descriptors. The 1D-CNN stands out with the highest accuracy of $97\%$ and a precision, recall, and F1 score of $0.97$. Compared to traditional machine learning models, this represents significant accuracy improvements: $15.13\%$ over LDA, $4.17\%$ over SVM, $3.34\%$ over KNN, $3.63\%$ over Random Forest, $2.88\%$ over Voting Ensemble, $2.83\%$ over Bagging KNN, $3.87\%$ over Bagging SVM, and $3.62\%$ over Adaboost. 1D Dilated CNN-LSTM model shows similar performance as that of 1D CNN with an accuracy of $96.53\%$.

Among traditional machine learning models, Voting Ensemble and Bagging KNN perform slightly better than individual models like SVM, KNN, and Random Forest. Specifically, Voting Ensemble shows a $0.74\%$ improvement over Random Forest, while Bagging KNN demonstrates a $0.78\%$ improvement over the same. These improvements illustrate the benefit of combining multiple models to improve robustness and reduce variance. The Support Vector Machine (SVM), K-Nearest Neighbors (KNN) and Random Forest classifiers perform similarly, with an accuracy of around $93-94\%$. Linear Discriminant Analysis (LDA), being a simpler model, shows the lowest performance with an accuracy of $84.26\%$.

\begin{table}[ht]
\centering
\caption{Gesture recognition from temporal-spatial descriptors-based feature extraction using traditional machine learning and deep learning tested on the Grabmyo dataset.}
\label{tab:TSDemg}
\resizebox{\columnwidth}{!}{%
\begin{tabular}{|l|c|c|c|c|}
\hline
\textbf{Models} & \textbf{ACC} & \textbf{P} & \textbf{R} & \textbf{F1} \\ \hline
LDA   & 88.1  &  0.89    & 0.88   &   0.88          \\
SVM   &   94.71 &   0.95   &  0.95  &  0.95         \\
KNN  &   94.94    &   0.95        &   0.95         &     0.95        \\
Random Forest  &   94.85  &   0.95   &   0.95   &   0.95       \\ 
Voting Ensemble  &      95.77      &   0.96         &   0.96         &     0.96       \\
Bagging KNN  &    95.04    &   0.95          &   0.95          &     0.95       \\
Bagging SVM  &      95.49   &   0.96          &   0.95          &     0.96        \\
Adaboost  &     95.04     &   0.95          &   0.95          &     0.95        \\
1D Dilated CNN  &      \textbf{ 96 }    &   0.97          &   0.96         &     0.96        \\
1D Dilated CNN-LSTM  &       95.97    &   0.96          &   0.96         &     0.96        \\
\hline
\end{tabular}%
}
\end{table}

Table~\ref{tab:TSDemg} presents the performance metrics for various machine learning and deep learning models applied to gesture recognition using temporal-spatial descriptors-based feature extraction.  1D dilated CNN stands out with the highest accuracy of $96\%$ and a precision, recall, and F1 score of $0.97$. Compared to traditional machine learning models, this represents significant accuracy improvements: $8.96\%$ on LDA, $1.36\%$ on SVM, $1.12\%$ on KNN, $1.21\%$ on Random Forest, $0.24\%$ on Voting Ensemble, $1.01\%$ over Bagging KNN, $0.53\%$ over Bagging SVM, and $1.01\%$ over Adaboost. 1D Dilated CNN-LSTM model shows similar performance as that of 1D CNN with an accuracy of $95.97\%$.

Among traditional machine learning models, the voting ensemble and Bagging KNN perform slightly better than individual models like SVM, KNN, and Random Forest. Specifically, Voting Ensemble shows a $0.97\%$ improvement over Random Forest, while Bagging KNN demonstrates a $0.20\%$ improvement over the same. These improvements illustrate the benefit of combining multiple models to enhance robustness and reduce variance. The Support Vector Machine (SVM), K-Nearest Neighbors (KNN) and Random Forest classifiers perform similarly, with an accuracy of around $94-95\%$. Linear Discriminant Analysis (LDA), being a simpler model, shows the lowest performance with an accuracy of $88.1\%$.

\begin{table}[ht]
\centering
\caption{Gesture recognition from Wavelet transform-based feature extraction using traditional machine learning and deep learning tested on the Grabmyo dataset.}
\label{tab:WTfeat}
\resizebox{\columnwidth}{!}{%
\begin{tabular}{|l|c|c|c|c|}
\hline
\textbf{Models} & \textbf{ACC} & \textbf{P} & \textbf{R} & \textbf{F1} \\ \hline
LDA   & 87.45  &  0.88    & 0.87   &   0.88          \\
SVM   &   92.36 &   0.92   &  0.92  &  0.92         \\
KNN  &   92.68    &   0.93        &   0.93         &     0.93        \\
Random Forest  &   92.78  &   0.93   &   0.93   &   0.93       \\ 
Voting Ensemble  &      94.18      &   0.94         &   0.94         &     0.94       \\
Bagging KNN  &   92.80     &   0.93          &   0.93          &     0.93       \\
Bagging SVM  &    93.46   &   0.94   &   0.93          &     0.93        \\
Adaboost  &      92.90    &   0.93    &   0.93    &  0.93        \\
1D Dilated CNN  &      \textbf{ 96 }    &   0.97          &   0.96         &     0.96        \\
1D Dilated CNN-LSTM  &      95.19    &   0.95          &   0.95         &     0.95        \\
\hline
\end{tabular}%
}
\end{table}

Table~\ref{tab:WTfeat} presents the performance metrics for various machine learning and deep learning models applied to gesture recognition using Wavelet Transform-based feature extraction. The 1D Dilated CNN stands out with the highest accuracy of $96\%$, and a precision, recall, and F1 score all of $0.97$. Compared to traditional machine learning models, this represents significant accuracy improvements: $9.78\%$ over LDA, $3.93\%$ over SVM, $3.58\%$ over KNN, $3.47\%$ over Random Forest, $1.93\%$ over Voting Ensemble, $3.45\%$ over Bagging KNN, $2.71\%$ over Bagging SVM, and $3.34\%$ over Adaboost. 1D Dilated CNN-LSTM model shows comparable performance as that of 1D CNN with an accuracy of  $95.19\%$.

Among traditional machine learning models, the voting ensemble and the bagging SVM perform slightly better than individual models such as SVM, KNN, and Random Forest. Specifically, Voting Ensemble shows a $1.51\%$ improvement over Random Forest, while Bagging SVM demonstrates a $0.73\%$ improvement over the same. These improvements illustrate the benefit of combining multiple models to improve robustness and reduce variance. The Support Vector Machine (SVM), K-Nearest Neighbors (KNN) and Random Forest classifiers perform similarly, with an accuracy of around $92-93\%$. Linear Discriminant Analysis (LDA), being a simpler model, shows the lowest performance with an accuracy of $87.45\%$.

In \textbf{summary}, Ensemble learning, 1D Dilated CNN and 1D Dilated CNN-LSTM  performed best in terms of precision when evaluated on the Grabmyo dataset. Among all features and models; with fused time-domain descriptors, 1D Dilated CNN outperformed SVM by $4.17\%$ and LDA by $15.13\%$, achieving an improved accuracy of $97\%$ on the Grabmyo dataset.

In \textit{comparison to the existing work}~\cite{kok2024machine} that uses Grabmyo dataset with only five basic hand gestures and trained LDA and SVM with time-domain features and obtaining an accuracy of $90.69\%$, our proposed work (1D Dilated CNN) uses the Grabmyo dataset with all gesture classes and exhibits a performance improvement of $6.96\%$ over this existing work.

\subsection{Experimental Results for FORS-EMG}

\begin{table}[ht]
\centering
\caption{Gesture recognition from fused time-domain descriptors using traditional machine learning and deep learning tested on FORS-EMG dataset.}
\label{tab:TDDemgFORS}
\resizebox{\columnwidth}{!}{%
\begin{tabular}{|l|c|c|c|c|}
\hline
\textbf{Models} & \textbf{ACC} & \textbf{P} & \textbf{R} & \textbf{F1} \\ \hline
LDA   & 62.18  &  0.62    & 0.62   &   0.62          \\
SVM   &   71.11 &   0.71   &  0.71  &  0.71         \\
KNN  &   92.61    &   0.93        &   0.93         &     0.93        \\
Random Forest  &   \textbf{93.85}  &   0.94   &   0.94   &   0.94        \\ 
Voting Ensemble  &       91.94      &   0.92          &   0.92         &     0.92       \\
Bagging KNN  &       92.9    &   0.93          &   0.93          &     0.93        \\
Bagging SVM  &     71.03     &   0.71          &   0.71          &     0.71        \\
Adaboost  &      \textbf{93.71}   &   0.94          &   0.94          &     0.94        \\
1D Dilated CNN  &      92.58     &   0.93          &   0.93         &     0.93        \\
1D Dilated CNN-LSTM  &      93.14     &   0.93          &   0.93         &     0.93        \\
\hline
\end{tabular}%
}
\end{table}

Table~\ref{tab:TDDemgFORS} presents the performance metrics for various machine learning and deep learning models applied to gesture recognition using fused time domain descriptors. Random Forest obtained an accuracy of $93.85\%$, representing a significant improvement of $50.84\%$ compared to LDA and $31.96\%$ compared to SVM. Similarly, the Voting Ensemble model obtained an accuracy of $91.94$, showing improvements of $47.81\%$ over LDA and $29.32\%$ over SVM. The 1D Dilated CNN, a deep learning model, obtained an accuracy of $92.58\%$, with a $48.85\%$ improvement compared to LDA and a $30.47\%$ improvement over SVM. In general, the Random Forest and Voting Ensemble models outperformed traditional machine learning methods, while the 1D Dilated CNN and 1D Dilated CNN-LSTM demonstrated competitive performance, showing a notable improvement in accuracy over both LDA and SVM.

\begin{table}[ht]
\centering
\caption{Gesture recognition from temporal-spatial descriptors-based feature extraction using traditional machine learning and deep learning tested on FORS-EMG Dataset.}
\label{tab:TSDemgFORS}
\resizebox{\columnwidth}{!}{%
\begin{tabular}{|l|c|c|c|c|}
\hline
\textbf{Models} & \textbf{ACC} & \textbf{P} & \textbf{R} & \textbf{F1} \\ \hline
LDA   & 63.3  &  0.64    & 0.63   &   0.63          \\
SVM   &   72.65 &   0.73   &  0.73  &  0.73         \\
KNN  &   89.9    &   0.90        &   0.90         &     0.90        \\
Random Forest  &   \textbf{94.95}  &   0.95   &   0.95   &   0.95       \\ 
Voting Ensemble  &       90.38     &   0.90         &   0.90         &     0.90       \\
Bagging KNN  &    90.34    &   0.90          &   0.90          &     0.90      \\
Bagging SVM  &      72.6   &   0.73          &   0.73          &     0.73        \\
Adaboost  &     \textbf{94.95}     &   0.95          &   0.95          &     0.95        \\
1D Dilated CNN  &      91.44    &   0.91          &   0.91         &     0.91        \\
1D Dilated CNN-LSTM  &      92.48    &   0.92          &   0.92         &     0.92        \\
\hline
\end{tabular}%
}
\end{table}

From Table~\ref{tab:TSDemgFORS}, we can observe the performance of various models for gesture recognition using temporal-spatial descriptors-based feature extraction. Among traditional machine learning models, K-Nearest Neighbors (KNN) obtained the highest accuracy of $89.05\%$, with precision (P), recall (R), and F1-score all of $0.89$. Compared to other traditional methods like LDA, SVM, and KNN showed substantial improvements. Specifically, compared to LDA ($63.3\%$), KNN improved accuracy by $40.66\%$, and compared to SVM ($72.65\%$), the improvement was $22.62\%$.

Random Forest model obtained an accuracy of $94.95\%$, which represents a $49.97\%$ improvement over LDA and a $30.74\%$ improvement over SVM. Similarly, the Voting Ensemble method, with an accuracy of $90.38\%$, outperformed LDA and SVM with an improvement of $42.74\%$ and $24.37\%$, respectively. Ensemble methods like Bagging KNN and Bagging SVM also performed well, with Bagging KNN obtaining $90.34\%$ accuracy, showing a $42.72\%$ improvement over LDA and a $24.29\%$ improvement over SVM. Adaboost demonstrated even higher accuracy at $94.95\%$, matching Random Forest's performance and showing a $49.97\%$ improvement over LDA and a $30.74\%$ improvement over SVM.

Among deep learning models, the 1D Dilated CNN achieved $91.44\%$ accuracy, showing a $44.46\%$ improvement over LDA and a $25.83\%$ improvement over SVM. This model also had consistent precision, recall, and F1-score values of $0.91$. Similarly, 1D Dilated CNN-LSTM achieved $92.48\%$ accuracy, showing performance improvement over LDA and SVM.

\begin{table}[ht]
\centering
\caption{Gesture recognition from wavelet transform-based feature extraction using traditional machine learning and deep learning tested on FORS-EMG dataset .}
\label{tab:WTfeatFORS}
\resizebox{\columnwidth}{!}{%
\begin{tabular}{|l|c|c|c|c|}
\hline
\textbf{Models} & \textbf{ACC} & \textbf{P} & \textbf{R} & \textbf{F1} \\ \hline
LDA   & 63.02  &  0.63    & 0.63   &   0.63          \\
SVM   &   72.13 &   0.72   &  0.72  &  0.72         \\
KNN  &   \textbf{94.41}    &   0.94        &   0.94         &     0.94        \\
Random Forest  &   92.8  &   0.93   &   0.93   &   0.93       \\ 
Voting Ensemble  &       92.36     &   0.92         &   0.92         &     0.92       \\
Bagging KNN  &   \textbf{94.52}    &   0.95          &   0.95          &     0.95       \\
Bagging SVM  &    72.52   &   0.73   &   0.73          &     0.73        \\
Adaboost  &      92.72    &   0.93    &   0.93    &  0.93        \\
1D Dilated CNN  &      92.85     &   0.93          &   0.93         &     0.93        \\
1D Dilated CNN-LSTM  &      93.05    &   0.93          &   0.93         &     0.93        \\
\hline
\end{tabular}%
}
\end{table}

In Table~\ref{tab:WTfeatFORS}, we observe the performance of various models for gesture recognition using wavelet transform-based feature extraction. Among traditional machine learning models, K-Nearest Neighbors (KNN) obtained the highest accuracy of $94.41\%$, with precision (P), recall (R), and F1-score all at $0.94$. Compared to other traditional models like LDA and SVM, KNN showed substantial improvements. Specifically, compared to LDA ($63.03\%$), KNN improved accuracy by $49.81\%$, and compared to SVM ($72.13\%$), the improvement was $30.95\%$.
Random Forest model achieved an accuracy of $92.8\%$, which represents a $47.22\%$ improvement over LDA and a $28.68\%$ improvement over SVM. Similarly, the Voting Ensemble method, with an accuracy of $92.36\%$, outperformed LDA and SVM with improvements of $46.55\%$ and $28.06\%$, respectively.

Ensemble methods like Bagging KNN and Bagging SVM also performed well, with Bagging KNN achieving $94.52\%$ accuracy, showing a $50.00\%$ improvement over LDA and a $31.06\%$ improvement over SVM. Adaboost demonstrated an accuracy of $92.72\%$, showing a $47.13\%$ improvement over LDA and a $28.57\%$ improvement over SVM. Among deep learning models, the 1D Dilated CNN obtained $92.85\%$ accuracy, showing a $47.31\%$ improvement over LDA and a $28.71\%$ improvement over SVM. This model also had a consistent precision, recall, and F1 score of $0.93$. Similarly, 1D Dilated CNN-LSTM obtained accuracy of $93.05\%$ showing improvement over LDA and SVM.

In \textbf{summary}, with fused time-domain descriptors, 1D Dilated CNN obtained an enhanced accuracy of $97\%$ on the Grabmyo dataset, outperforming SVM by $4.17\%$ and LDA by $15.13\%$. Similarly with temporal feature descriptors, the Random Forest~(RF) model obtained an accuracy of $94.95\%$ for the FORS-EMG dataset, outperforming SVM by
$30.74\%$ and LDA by $49.97\%$.

\section{Conclusions}

This work benchmarks machine learning and deep learning models trained with several feature extraction techniques, including wavelet transform-based features, temporal-spatial descriptors, and time-domain descriptors, to test the performance of EMG-based gesture recognition. Two publicly available datasets were tested, namely Grabmyo and FORS-EMG. For Grabmyo among all feature extraction and ML/DL methods; 1D dilated CNN outperformed SVM by $4.17\%$ and LDA by $15.13\%$, obtaining an impressive accuracy of $97\%$ with time-domain features. Similarly, the FORS-EMG dataset among all feature extraction and ML/DL methods; the Random Forest model obtained an accuracy of $94.95\%$, which represents a $49.97\%$ improvement over LDA and a $30.74\%$ improvement over SVM when used with temporal feature descriptors. As a part of future work, advanced deep learning models will be implemented to further improve hand-based gesture recognition performance.


\small{
\flushend
\bibliographystyle{IEEE}
\bibliography{conference_101719}}
\end{document}